\documentclass[journal, 10pt, twoside]{IEEEtran}

\usepackage{amsmath,amssymb,amsfonts}
\usepackage{algorithmic}
\usepackage{graphicx}
\usepackage{textcomp}
\usepackage{xcolor}
\usepackage[caption=false]{subfig}

\usepackage[backend=biber,bibstyle=ieee,citestyle=numeric,sorting=none,sortcites=true]{biblatex}

\usepackage[hidelinks]{hyperref}
\usepackage{orcidlink}
\usepackage{acronym}
\usepackage[nameinlink, noabbrev]{cleveref}
\usepackage{csquotes}
\usepackage[group-separator={\,},
            group-minimum-digits=5,
            group-digits=integer,
            uncertainty-mode=separate,
            retain-zero-uncertainty=true,
            ]{siunitx}
\usepackage{mathtools}
\usepackage{xspace}
\usepackage{booktabs}
\usepackage{multirow}
\usepackage[inline]{enumitem}
\usepackage{microtype}
\usepackage{fmtcount}

\usepackage{tikz}
\usetikzlibrary{calc, math, arrows.meta, positioning, fit, shapes.geometric, intersections}
\tikzset{
    nosep/.style={inner sep=0pt, outer sep=0pt},
    img/.style={nosep},
    cnn/.style={draw, trapezium, shape border rotate=270, text width=1.1cm, align=center, trapezium angle=70},
    tra/.style={draw, rounded corners=3pt, text width=1.1cm, align=center},
    pretrained/.style={fill=violet!25},
    arr/.style={-latex},
    lbl/.style={text width=#1, align=center, font=\small},
    lbl/.default={1.5cm},
}

\let\cite=\autocite
\newcommand{\model}[1]{\textit{#1}}
\newcommand{\dataset}[1]{\textit{#1}}
\newcommand{\software}[1]{\texttt{#1}}

\newcommand{\TDSRDH}{\model{TDSRDH}\xspace}
\newcommand{\CLIPHash}{\model{CLIPHash}\xspace}
\newcommand{\NUSWIDE}{\dataset{NUS-WIDE}\xspace}
\newcommand{\MIRFLICKR}{\dataset{\mbox{MIR Flickr 25k}}\xspace}

\newcommand{\modality}[1]{{(\text{#1})}}
\newcommand{\mmodality}[1]{{(#1)}}
\newcommand{\img}{\modality{I}}
\newcommand{\txt}{\modality{T}}
\newcommand{\itoi}{$\text{I}\kern-2pt\to\kern-1pt\text{I}$}
\newcommand{\itot}{$\text{I}\kern-2pt\to\kern-2pt\text{T}$}
\newcommand{\ttoi}{$\text{T}\kern-3pt\to\kern-1pt\text{I}$}
\newcommand{\ttot}{$\text{T}\kern-3pt\to\kern-2pt\text{T}$}

\DeclareMathOperator{\sign}{sign}
\DeclareMathOperator{\relu}{ReLU}

\usepackage{ifthen}
\usepackage{calc}
\newcommand{\mlpcfg}[2][]{%
    \newif\ifarrowprinted\arrowprintedfalse%
    $\foreach \d [count=\i from 0] in {#2}{%
        \arrowprintedfalse%
        \foreach \a/\b [count=\j from 1] in {#1} {%
            \ifnum\i=\j\relax%
                \ifthenelse{\equal{\a}{\b}}{\def\b{}}{}%
                \xrightarrow[{\raisebox{1.25ex-\heightof{$\scriptstyle\b$}}[0pt]{$\scriptstyle\b$}}]{\a}%
                \global\arrowprintedtrue\breakforeach%
            \fi%
        }%
        \ifarrowprinted\else\ifnum\i>0\relax\xrightarrow{}\fi\fi%
        \d%
    }$%
}

\begin{document}
\acrodef{ap}[AP]{Average Precision}
\acrodef{bow}[BoW]{Bag of Words}
\acrodef{cnn}[CNN]{Convolutional Neural Network}
\acrodef{dsh}[DSH]{Deep Semantic Hashing}
\acrodef{dsch}[DSCH]{Dynamic Semantic Channel Hashing}
\acrodef{map}[mAP]{Mean Average Precision}
\acrodef{mlp}[MLP]{Multilayer Perceptron}
\acrodef{nlp}[NLP]{Natural Language Processing}
\acrodef{sch}[SCH]{Semantic Channel Hashing}

\def\fulltitle{DSCH-Loss: A Dynamic Semantic Channel Objective for Deep Semantic Hashing}
\title{\fulltitle}

\author{%
    \hfill
    \IEEEauthorblockN{Tobias J. Bauer\textsuperscript{a} \orcidlink{0009-0006-1073-3971}}
    \hfill
    \IEEEauthorblockN{Christian Riess\textsuperscript{b} \orcidlink{0000-0002-5556-5338}}
    \hfill
    \IEEEauthorblockN{Daniel Loebenberger\textsuperscript{a,c} \orcidlink{0000-0002-7969-6260}}
    \hfill
    \IEEEauthorblockN{Christian Bergler\textsuperscript{c} \orcidlink{0009-0005-1793-5798}}
    \hfill\!\\[1mm]
    \begin{minipage}[t]{0.425\linewidth}
        \IEEEauthorblockA{\centering%
            \textsuperscript{a}\textit{Fraunhofer Institute for Applied and\\Integrated Security AISEC}\\
            Weiden, Germany \\
            {\tt\hskip-1.5em \{tobias.bauer, daniel.loebenberger\}\kern-2em\\@aisec.fraunhofer.de}}
    \end{minipage}%
    \begin{minipage}[t]{0.3\linewidth}
        \IEEEauthorblockA{\centering%
            \textsuperscript{b}\textit{Friedrich-Alexander-Universität Erlangen-Nürnberg}\\
            Erlangen, Germany \\
            {\tt christian.riess\\@fau.de}}
    \end{minipage}%
    \begin{minipage}[t]{0.275\linewidth}
        \IEEEauthorblockA{\centering%
            \textsuperscript{c}\textit{Ostbayerische Technische\\Hochschule Amberg-Weiden}\\
            Amberg, Germany \\
            {\tt c.bergler\\@oth-aw.de}}
    \end{minipage}%
}
\markboth{}{Bauer \MakeLowercase{\textit{et al.}}: \fulltitle}

\maketitle

\begin{abstract}\sisetup{text-series-to-math}%
Semantic hashing methods for generating short binary hash codes that allow efficient approximate nearest neighbor search in high-dimensional data spaces have gained extensive consideration in recent years. Deep learning-based methods offer better semantic capturing capabilities than traditional approaches relying on manual feature engineering. Moreover, they enable a data-driven approach to semantic hashing across diverse data modalities, yielding high-quality cross-modal hash codes within a shared Hamming space. Previous work investigated the properties of this Hamming space and introduced a loss function based on predefined so-called semantic channels with fixed width and Hamming distances derived from label similarities. However, this formulation also introduced discontinuities into the loss landscape, complicating optimization. Based on these observations, we propose a newly designed loss function, \acf{dsch}, using dynamically sized and positioned semantic channels in order to avoid loss landscape discontinuities. Furthermore, we endorse the use of tie-aware \acf{map} as evaluation metric as it addresses the ambiguity in sample retrieval ordering, which emerges from the discreteness of hash code distances. Finally, multiple experimental settings conducted on two popular datasets and incorporating two different model architectures provide strong evidence that training using the \ac{dsch} objective outperforms training using other state-of-the-art loss functions. In a total of \num{35} out of \num{40} cross-modal and intra-modal retrieval tasks, models trained with \ac{dsch} achieve significantly higher tie-aware \ac{map} scores across all four tested hash code lengths, showing compelling results across model architecture and used dataset. The \ac{map} score uplifts are consistent and amount up to \num{1.75} percentage points compared to the respective second best. \acresetall
\end{abstract}

\begin{IEEEkeywords}
Deep semantic hashing, Cross-modal retrieval, Hamming space, Deep learning
\end{IEEEkeywords}

\section{Introduction}\label{s:intro}
\IEEEPARstart{T}{he} task of finding approximate nearest neighbors in high-dimensional data spaces can be tackled using \ac{dsh} as the application of deep neural networks to the field of Semantic Hashing~\cite{Luo2023,Singh2022}. Given a query sample, the goal is to retrieve semantically similar samples from a database or retrieval set. Mapping data samples to compact binary hash codes, i.e., hashing the samples, is a viable and storage-efficient~\cite{Luo2023} way to represent the semantics. In general, a hash function takes data of arbitrary length -- often represented as a bit-string of any length -- and outputs a binary hash code of a given fixed length, e.g., \num{32}\,bits. These short hash codes allow for computationally efficient nearest-neighbor search~\cite{Luo2023,Singh2022}. Furthermore, \ac{dsh} is also well-suited for cross-modal retrieval which extends the single-modal retrieval setting with multi-modal data samples and the task of finding similar samples in a different modality~\cite{Wang2024,Luo2023,Zhou2023a}. This is due to its data-driven training approach using multi-modal samples sharing a common semantic.

The remainder of this paper is organized as follows. \Cref{s:rw} presents fundamental concepts of \ac{dsh} and cross-modal retrieval as well as specific model architectures and training procedures used as a basis for our experiments. In \Cref{s:loss}, we propose a new and novel loss function for \ac{dsh}. Next, \Cref{s:exp} describes the experimental setup, including the data material and model architectures as well as data preprocessing procedures, the training protocol, and evaluation metrics. Further, the experimental results are presented and analyzed in \Cref{s:res}. \Cref{s:conc} concludes this paper and provides perspectives on future research.

\section{Related Work}\label{s:rw}
\subsection{(Deep) Semantic Hashing}\label{s:rw:dsh}
Hashing is the process of mapping high-dimensional data to short binary hash codes by projecting the input into a low-dimensional Hamming space~\cite{Luo2023}. Such high-dimensional input data may comprise various modalities, including images, text, audio, and video, as well as other structured or unstructured data types~\cite{Wang2024,Luo2023,Singh2022}. In the context of Semantic Hashing, the goal is to find a hash function that maps semantically similar inputs to similar or identical hash codes to facilitate nearest neighbor search~\cite{Luo2023,Singh2022}. The  Hamming distance between any two hash codes can be easily computed as the number of differing binary digits (bits). Good semantic hash functions are characterized by their ability to output hash codes that are close, i.e., have a small distance, for similar inputs and simultaneously output distant hash codes for dissimilar inputs~\cite{Luo2023}. There are different algorithmic goals a concrete semantic hash function may pursue, e.g., it may be optimized for creating hash tables or it may be designed for hash code ranking~\cite{Wang2024}. Either way, these hash functions can either be crafted or learned and the paradigm of \acl{dsh} emerges from end-to-end hash function learning using deep neural networks~\cite{Luo2023}. \citeauthor{Luo2023}~\cite{Luo2023} present a comprehensive survey on this topic.

\subsection{Cross-Modal Retrieval}\label{s:rw:cm}
As stated in the previous \namecref{s:rw:dsh}, Semantic Hashing allows for an efficient similarity search using high-quality hash codes as surrogates for the high-dimensional underlying data. In previous work, the hash function operates on single-modal input data, e.g., only images or only text, see~\cite{Lu2017,Li2020,Wang2021,Chen2022,Xuan2021}, with one of the earliest work in the cross-modal \ac{dsh} domain done by \citeauthor{Jiang2017}~\cite{Jiang2017}.

More recent research, however, focuses on multi-modal hashing, which aims to project data from different modalities into a shared Hamming space~\cite{Luo2023,Wang2024}. This allows for so-called cross-modal retrieval, such that for example images can be searched using text and vice versa. In their literature review, \citeauthor{Wang2024}~\cite{Wang2024} present an elaborate taxonomy on cross-modal retrieval, differentiating between supervised and unsupervised as well as real-valued and hashing-based approaches. Although many combinations of media modalities are studied, the main focus of research currently lies upon the text-image retrieval~\cite{Wang2024} as seen in~\cite{Hong2022,Li2023,Jin2023,Zhou2023,Xia2023,Huang2024,Li2024,Hu2024,Zhu2024}.\vspace{-1.5ex}

\subsection{Model Architectures}\label{s:rw:models}
In the context of cross-modal \ac{dsh} a variety of neural network architectures have been proposed and applied to the problem~\cite{Luo2023,Wang2024}. A fundamental approach is to use an off-the-shelf and often pre-trained image model, e.g., a \ac{cnn}~\cite{LeCun2010,Krizhevsky2012} such as ResNet~\cite{He2015}, and replace its classification head with a hash-learning \ac{mlp}~\cite{Luo2023,Wang2024}. For the text modality, it is common to use simple \acp{mlp} projecting \ac{bow} vectors to hash codes as seen in~\cite{Jiang2017,Deng2018,Mingyong2023,Hu2024}. However, recent publications incorporate the Transformer architecture~\cite{Vaswani2017} that operates on token embeddings and which underpins recent advances in \acl{nlp}~\cite{Supriyono2024}. \Ac{dsh} model architectures can be categorized into two primary approaches: single-stream and dual-stream architectures~\cite{Zhou2023a,Wang2024}. The former consists of a single module that operates on inputs of different modalities in a uniform manner, whereas the latter employs one module per modality~\cite{Zhou2023a,Wang2024}. There also exist hybrid approaches, which combine per-modality feature extraction with a shared hash-learning module~\cite{Zhou2023a}.\vspace{-1.5ex}

\subsection{\acl{sch}}\label{s:rw:sch}
\citeauthor{Hu2024}~\cite{Hu2024} report a compression of the Hamming space when using common \ac{dsh} loss functions which constrain dissimilar samples to be either orthogonal or as distant as possible in Hamming space. In the first case, the space of fully or partially similar samples gets compressed and almost half of the Hamming space is not used according to the authors. In the latter case, semantically similar samples are able to utilize more of the Hamming space, however, now the space for dissimilar samples gets compressed, which reduces the separability of these samples~\cite{Hu2024}.

To overcome the limitations of current loss function constraints, the authors propose a novel loss function named \acf{sch}~\cite{Hu2024}. They address the Hamming space compression issue by introducing target channels for pairs of samples. The \ac{sch} loss ensures that during training the model is punished for putting pairs of samples too close or too distant within the Hamming space and endures no penalties when the Hamming distance lies within the target channel. These channels have a fixed width of $\tau$ and their position within the Hamming space is determined by the label-similarity of the samples~\cite{Hu2024}. As their work provides the theoretical foundation for our loss formulation, the concept of semantic channels is elaborated in detail in \Cref{s:loss:channels}.

The authors report state-of-the-art results using their method on two datasets~\cite{Hu2024}. Even though the used evaluation metric is widespread, the results are not directly comparable as the metric formula is not permutation-agnostic and, thus, produces different scores for semantically identical predictions. This issue is addressed in~\cite{He2018} and elaborated upon in \Cref{s:exp:eval}.

\section{Proposed Method}\label{s:loss}
\subsection{Problem Statement and Notation}\label{s:loss:problem}
We define the datasets to be comprised of $n$ samples, i.e., $\mathcal{D}=\{x_i\}_{i=0}^{n}$, with each sample a pair of image and text, i.e., $x_i = \langle x_i^\img, x_i^\txt \rangle$. Furthermore, each sample is annotated with a one-hot encoded label vector $l_i \in \{0, 1\}^c$ where $c$ is the number of labels in the dataset. The to be learned cross-modal hash function accepts inputs $x_i^\modality{M}$ of modality $\text{M}$, e.g., any image $x_i^\img$ or text $x_i^\txt$, and produces a $k$-bit hash code $b_i^\modality{M} \in \{-1, 1\}^k$ after applying the $\sign$ function to the model output $\hat{b}_i^\modality{M} \in \mathbb{R}^k$. The notation used throughout this paper is summarized in \Cref{t:notation}.

Both datasets used in this paper consider two samples similar if they have at least one label in common, mathematically $x_i \sim x_j \iff l_i \cdot l_j > 0$. Our goal, however, is to construct a loss function that incorporates real-valued semantic similarity of samples. The more the label vectors of two samples align, the higher their similarity score. Following~\cite{Hu2024}, we use the cosine similarity between label vectors as inter-sample similarity score:\vspace{-1mm}
\begin{equation}
    \mathbf{S}_{ij} = \cos\left(l_i, l_j\right) = \frac{l_i \cdot l_j}{\lvert l_i \rvert \cdot \lvert l_j \rvert} \in [0, 1],\quad 0 \leq i, j \leq n.
    \label{eq:sim}
\end{equation}
Furthermore, \citeauthor{Hu2024}~\cite{Hu2024} formulate the Hamming distance~$d$ between two binary hash codes using the cosine similarity:
\begin{equation}
    d\left(b_i^\mmodality{m_i}, b_j^\mmodality{m_j}\right)= \frac{k}{2}\left(1 - \cos\left(b_i^\mmodality{m_i}, b_j^\mmodality{m_j}\right)\right).
    \label{eq:dis}
\end{equation}\vspace{-2ex}

\begin{table}[htbp]
    \centering
    \caption{Mathematical notation.}\label{t:notation}
    \begin{tabular}{@{}l@{\hspace{5pt}}p{\dimexpr\linewidth-29mm}@{}}\toprule
         \textbf{Notation} & \textbf{Description}\\\midrule
         $\mathcal{D}$ & Particular dataset, e.g., $\mathcal{Q}$ query, $\mathcal{R}$ retrieval,  $\mathcal{T}$ train\\
         $n = \lvert \mathcal{D} \rvert$ & Number of samples in selected dataset $\mathcal{D}$ \\
         $x_i = \langle x_i^\modality{I}, x_i^\modality{T} \rangle$ & $i$-th sample pair in $\mathcal{D}$ comprising of image and text \\
         $c$ & Number of labels in $\mathcal{D}$,\hfill\break e.g., $c=21$ for \NUSWIDE (see~\Cref{s:exp:data}) \\
         $l_i \in \{0, 1\}^c$ & One-hot encoded labels of sample $x_i$ \\
         $k$ & Hash code length in bits \\
         $b_i^\modality{M} \in \{-1, 1\}^k$ & Binary hash code of sample $x_i^\modality{M}$ of modality $\text{M}$ \\
         $\hat{b}_i^\modality{M} \in \mathbb{R}^k$ & Real-valued hash code of $x_i^\modality{M}$ prior quantization\\
         $\mathbf{X}^\img \in \mathbb{R}^{n \times d_I}$ & Images of $\mathcal{D}$ with data dimensionality $d_I$\\
         $\mathbf{X}^\txt \in \mathbb{R}^{n \times d_T}$ & Texts of $\mathcal{D}$ with data dimensionality $d_T$\\
         $\mathbf{L} \in \{0, 1\}^{n \times c}$ & One-hot encoded labels of $\mathcal{D}$ \\
         $\mathbf{S} \in \left[0, 1\right]^{n \times n}$ & Inter-sample label cosine similarity \\
         $\mathbf{B}^\modality{M} \in \{-1, 1\}^{n \times k}$ & Binary hash codes of samples of modality $\text{M}$ in $\mathcal{D}$ \\
         $\mathbf{\hat{B}}^\modality{M} \in \mathbb{R}^{n \times k}$ & Real-valued hash codes prior quantization into $\mathbf{B}^\modality{M}$ \\\bottomrule
    \end{tabular}
\end{table}

\subsection{Semantic Channels}\label{s:loss:channels}
\citeauthor{Hu2024}~\cite{Hu2024} introduced the notion of semantic channels. Their goal was to \enquote{allocate [...] an appropriate [target] Hamming distance to the hash codes $b_i$ and $b_j$ of samples $x_i$ and $x_j$ [...] based on their similarity $\mathbf{S}_{ij}$,}~\cite{Hu2024} which is the cosine similarity of their labels as calculated in \Cref{eq:sim}. In essence, the authors assigned each pair of samples a $\tau$-wide \enquote{semantic channel} within the Hamming space. The notion of a semantic channel is geometrically equivalent to a hyperspherical shell or $k$-dimensional annulus. The position in terms of distance in the Hamming space of said channels is based on their samples' label similarities. The loss function is designed so that the model is incentivized to project two samples into the shared Hamming space in a manner that satisfies the target Hamming distance constraint.

In their paper, \citeauthor{Hu2024}~\cite{Hu2024} split pairs of samples into three distinct sets based on their label similarity: negative \mbox{($\mathbf{S}_{ij} = 0$)}, fully positive \mbox{($\mathbf{S}_{ij} = 1$)}, and partially positive \mbox{($0 < \mathbf{S}_{ij} < 1$)}. Next, they defined a compound loss function on these three sets, using the label similarity and \Cref{eq:dis} to calculate the upper and lower target Hamming distances $\lambda^\modality{u}$ and $\lambda^\modality{l}$, respectively. These upper and lower bounds define the aforementioned channel. We re-formulate the equations given in~\cite{Hu2024} into a harmonized and simplified set of equations:
\begin{align}
    \lambda_{ij}^\modality{l} &= \begin{cases*}
        \frac{k}{2}                                         & if $\mathbf{S}_{ij} = 0$, \\
        \frac{k}{2} \left(1 - \mathbf{S}_{ij}\right) - \tau & otherwise. \\
    \end{cases*}\label{eq:sch:lambdal} \\[2pt]
    \lambda_{ij}^\modality{u} &= \begin{cases*}
        k                                            & if $\mathbf{S}_{ij} = 0$, \\
        \frac{k}{2} \left(1 - \mathbf{S}_{ij}\right)\phantom{~- \tau} & otherwise. \\
    \end{cases*}\label{eq:sch:lambdau}
\end{align}
The channel width $\lambda_{ij}^\modality{u} - \lambda_{ij}^\modality{l}$ is equal to $\tau$ for partially similar or fully similar samples, whereas it is $\frac{k}{2}$ for dissimilar samples, i.e., $\left(\lambda_{ij}^\modality{u} - \lambda_{ij}^\modality{l}\right) \in \left\{\tau, \frac{k}{2}\right\}$. Analogously to \citeauthor{Hu2024}~\cite{Hu2024}, the loss function can be defined as a sum over all samples and all combinations of modalities:
\begin{gather}
\begin{aligned}
    \mathcal{L}_\text{SCH} &= \sum_{m_i, m_j}^{\{\text{I}, \text{T}\}} \sum_{i, j = 1}^{n} \psi_{ij} \max\left\{ 0, \lambda_{ij}^\modality{l} - d^*, d^* - \lambda_{ij}^\modality{u} \right\}, \\[5pt]
    \text{where~} d^* \hspace{-1.5pt} &= d\left(\hat{b}_i^\mmodality{m_i}, \hat{b}_j^\mmodality{m_j}\right) \text{the observed Hamming distance,} \\[5pt]
    \text{and~} \psi_{ij} &= \begin{cases*}
        \beta & if $\mathbf{S}_{ij} = 0$, \\
        \alpha & if $\mathbf{S}_{ij} = 1$, \\
        1 & otherwise \\
    \end{cases*} \text{~the similarity set weight}.
\end{aligned}\raisetag{\baselineskip}\label{eq:sch:loss}
\end{gather}
Following~\cite{Hu2024}, the weighting hyperparameters $\alpha$ and $\beta$ are set to \num{1}, effectively simplifying the weight in \Cref{eq:sch:loss} to $\psi_{ij}=1$ in all cases. However, even after this simplification, the similarity set distinctions in \Cref{eq:sch:lambdal,eq:sch:lambdau} still remain, introducing discontinuities in the loss landscape at $\mathbf{S}_{ij} = 0$. This is because the lower channel bound jumps from $\lambda_{ij}^\modality{l} = \frac{k}{2}$ for $\mathbf{S}_{ij} = 0$ to $\lambda_{ij}^\modality{l} \approx \frac{k}{2}-\tau$ for $\mathbf{S}_{ij} = \varepsilon$ with $\varepsilon>0$ a small positive number. Similarly, the upper channel bound jumps from $\lambda_{ij}^\modality{u} = k$ for $\mathbf{S}_{ij} = 0$ to $\lambda_{ij}^\modality{u} \approx \frac{k}{2}$ for $\mathbf{S}_{ij} = \varepsilon$. The loss landscape visualization of \ac{sch} depicted in \Cref{f:cmpll:sch} clearly shows the discontinuity. Addressing this loss function discontinuity is an integral part of our proposed loss described in \Cref{s:loss:proposed}.

In addition to the iterative loss formulation, \citeauthor{Hu2024} present a vectorized version~\cite[\nameCref{eq:sch:loss}~(10)]{Hu2024}. Albeit them stating it being an equivalent formulation~\cite{Hu2024}, there is a difference between the iterative and vectorized formulas in terms of Hamming distance calculation. In the iterative version they use cosine similarity as seen in \Cref{eq:dis}, whereas in the vectorized version they opted to approximate the Hamming distance by \mbox{$\mathbf{B}^\mmodality{m_i} \times \left(\mathbf{B}^\mmodality{m_j}\right)^T$}~\cite{Hu2024}. This is indeed equivalent in the case of truly binary hash codes, i.e., \mbox{$\mathbf{B}_{ij}^\modality{M} \in \{-1, 1\}$}, since for each hash code vector \mbox{$\lvert b_{i}^\modality{M}\rvert = \sqrt{k}$}. However, for backpropagation to be applicable, the loss function must be differentiable, which precludes the use of the $\sign$ function as it is non-differentiable at zero and its gradient is zero everywhere else. Therefore, the model outputs \enquote{almost-binary} hash codes with \enquote{bits} that are still real-valued, i.e., \mbox{$\mathbf{\hat{B}}_{ij}^\modality{M} \in \mathbb{R}$}. Consequently, the aforementioned Hamming distance approximation breaks since in general \mbox{$\lvert \hat{b}_{i}^\modality{M} \rvert \ne \sqrt{k}$}. For this reason, our vectorized formulation employs the cosine similarity for Hamming distance calculation as well~(see~\Cref{s:loss:proposed}).

Additionally to these findings, \citeauthor{Hu2024} introduce two Frobenius norms to their vectorized loss formulation. They claim better balancing between positive and negative samples~\cite{Hu2024}. However, we empirically found that incorporating the Frobenius norm regularization induces model collapse during training, yielding trivial solutions in which the network outputs identical hash codes regardless of input. Consequently, in~\Cref{s:loss:proposed} we defaulted to the \mbox{$\text{L}_{1,1}$-norm}, effectively summing all error terms, which is equivalent to the iterative formulation.

\subsection{Proposed Loss for \acl{dsch}}\label{s:loss:proposed}
In this \namecref{s:loss:proposed}, we present our proposed loss function for \acf{dsch} which is based on the \ac{sch} loss by \citeauthor{Hu2024}~\cite{Hu2024}. First, we specify the design goals that led to the formulation of the \ac{dsch} objective. Next, we give an iterative formulation as well as an optimized vectorized version. Then, we introduce an additional quantization loss to accommodate for different model output behaviors (see~\Cref{s:exp:model}). In a final step, all loss terms are combined into a joint total loss function. All hyperparameters used in the formulas are summarized in \Cref{t:hyperparams}.

\subsubsection{Design Goals} A major design goal for \ac{dsch} is to overcome the loss landscape discontinuity for slightly similar sample pairs, i.e., $\mathbf{S}_{ij} \approx 0$ as described in \Cref{s:loss:channels}. Likewise, this discontinuity also involves the width of the target semantic channel, which also depends on whether the similarity is zero or not. We, therefore, propose a dynamic channel width $w$ that is dependent on the label similarity and ranges from a configurable minimum channel width $\tau$ to $k - \lambda_\text{neg}$. For dissimilar sample pairs, we introduce a configurable minimum distance $\lambda_\text{neg} = \frac{k}{2}$, which is analogous to the behavior of \ac{sch}~\cite{Hu2024} (see~\Cref{eq:sch:lambdal}). It is important to note that the channel width scaling may be non-linear in order to enforce $\tau$-sized channels for almost similar sample pairs, i.e., $\mathbf{S}_{ij} \approx 1$. Additionally, our loss should incentivize the model to pull fully similar sample pairs, i.e., $\mathbf{S}_{ij} = 1$, close in Hamming space and, thus, should penalize any Hamming distance greater than zero for said pairs. Finally, as reasoned in \Cref{s:loss:channels}, a configurable loss curve hyperparameter is used instead of the Frobenius norm, with the default being a $\text{L}_{1,1}$-norm.

\subsubsection{Iterative Formulation} Our proposed loss uses dynamic channel widths for sample pairs based on their label cosine similarity. The following \Cref{eq:dsch:it:w,eq:dsch:it:p,eq:dsch:it:l} define the computation of the semantic channel width $w_{ij}$, the leftmost semantic channel point $p_{ij}$, and the proposed loss $\mathcal{L}_\text{D}$:
\begin{align}
    w_{ij} &= \left(1 - \mathbf{S}_{ij}\right)^{\gamma_w} \cdot \left(k - \lambda_\text{neg} - \tau\right) + \tau. \label{eq:dsch:it:w}\\
    p_{ij} &= \lambda_\text{neg} \cdot \left(1 - \mathbf{S}_{ij}\right) - \mathbf{S}_{ij} \cdot \tau. \label{eq:dsch:it:p}
\end{align}
\begin{gather}
\begin{aligned}
    \mathcal{L}_\text{D} = \kern1.3em&\kern-1.3em\kern-3pt \sum_{m_i, m_j}^{\{I, T\}} \sum_{i, j = 1}^n \left(\psi_{ij} \max\left\{0, p_{ij} - d^*, d^* - \left(p_{ij} + w_{ij}\right)\right\}\right)^{\gamma_\ell}, \\[5pt]
    \text{where~} d^* \hspace{-1.5pt} &= d\left(\hat{b}_i^\mmodality{m_i}, \hat{b}_j^\mmodality{m_j}\right) \text{the observed Hamming distance,}\\[5pt]
    \text{and~} \psi_{ij} &= \begin{cases*}
        \beta & if $\mathbf{S}_{ij} = 0$, \\
        \alpha & if $\mathbf{S}_{ij} = 1$, \\
        1 & otherwise \\
    \end{cases*} \text{~the similarity set weight}.
\end{aligned}\raisetag{\baselineskip} \label{eq:dsch:it:l}
\end{gather}
Following \Cref{eq:dsch:it:w}, the channel width gets interpolated between the minimum channel width $\tau$ and the reserved space for dissimilar samples, i.e., $w_{ij} \in [\tau, k- \lambda_\text{neg}]$. However, the interpolation is not linear, but controlled by the channel width curve modifier $\gamma_w$. This measure ensures that the width of the target semantic channel stays small for an extended period of similarity values while it approaches dissimilarity, and also retrains the continuity property. This is in stark contrast to the proposal of \ac{sch} by \citeauthor{Hu2024}~\cite{Hu2024} that does not interpolate the channel width, but jumps between two values as described in \Cref{s:loss:channels}. \Cref{f:cmpll} depicts the loss landscapes of both algorithms as per \Cref{eq:sch:loss,eq:dsch:it:l}. The target semantic channel forms a trench in both loss landscapes. Apart from this, the loss landscape surface of \ac{dsch} is smooth without discontinuities as visualized in \Cref{f:cmpll:dsch}. We also provide an interactive 2D-sliced comparison visualization\rlap{.}\footnote{\url{https://www.desmos.com/calculator/qog7tklzdz}}\vspace{-1ex}

\newcommand{\losslandscape}[4][0.5\linewidth]{%
\captionsetup{margin=6pt, hangindent=\widthof{(a)\,}, justification=raggedright}%
    \subfloat[#3]{%
        \begin{minipage}{#1}%
            \centering%
            \begin{tikzpicture}[inner sep=0pt, outer sep=0pt]
                \node (p) {\includegraphics[trim={2mm 2mm 0mm 5mm}, clip]{ll-#2-bg.pdf}};
                \node {\includegraphics[trim={2mm 2mm 0mm 5mm}, clip]{ll-#2-fg.png}};
                \begin{scope}[every node/.append style={font=\scriptsize, align=center}]
                    \path let \p1=(p.south west), \p2=(p.north east), \n1={scalar(\x2 - \x1)*0.03528}, \n2={scalar(\y2 - \y1)*0.03528}
                    in \pgfextra{\pgftransformshift{\pgfpoint{\x1}{\y1}}\pgftransformxscale{\n1}\pgftransformyscale{\n2}}
                    node (lz) at (0.03, 0.45) [rotate=90, text width=0.5\linewidth, anchor=south] {Loss value $\mathcal{L}$}
                    node (lx) at (0.31, 0.02) [rotate=-5, text width=0.5\linewidth, anchor=north] {Observed Hamming\\distance $d^*$}
                    node (ly) at (0.84, 0.19) [rotate=50, text width=0.4\linewidth, anchor=north] {Label similarity\\~$\mathbf{S}_{ij}$}
                    ;
                \end{scope}
                \pgfresetboundingbox\useasboundingbox node [fit=(p)(lz)(lx)] {};
            \end{tikzpicture}
        \end{minipage}%
        \label{#4}%
    }%
}
\begin{figure}[b]
    \centering\vspace{-3ex}%
    \losslandscape{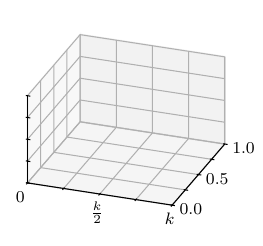}{\acl{sch} with $\tau=3$}{f:cmpll:sch}\hfill%
    \losslandscape{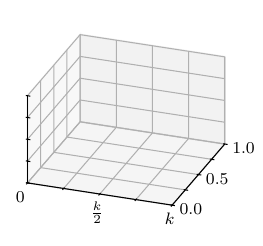}{\acl{dsch} with $\gamma_w=2, \tau=1$}{f:cmpll:dsch}%
    \caption{Visual comparison of the loss landscapes of \acl{sch} and \acl{dsch}. The loss landscape discontinuity at $\mathbf{S}_{ij} \approx 0$ of \ac{sch} is clearly visible, whereas \ac{dsch} boasts a smooth surface.}%
    \label{f:cmpll}%
\end{figure}

Besides continuity, there is another reason in favor of channel width interpolation: a low similarity score of two samples means that they are only slightly similar, which in turn carries a large uncertainty about the semantic basis of their similarity. Thus, a wider semantic channel accommodates this uncertainty, whereas a fixed-sized channel width would impose an unjustified constraint on the Hamming space. Conversely, a high label similarity score implies a clear and unambiguous basis of similarity, allowing a more restrictive semantic channel width. In this regime the uncertainty is low, thus, a narrower channel is sufficient and even preferable. This property allows decreasing the minimum channel width to \mbox{$\tau = 1$} compared to the relatively wide channel of~\num{3} in~\cite{Hu2024}. The semantic channel provides a distance-based ordering that guides the model at projecting the samples into the shared Hamming space. These insights provide compelling evidence in support of our proposal of non-linear channel width interpolation.

The leftmost semantic channel point $p_{ij}$ depends linearly on the label cosine similarity as stated in \Cref{eq:dsch:it:p}. This is analogous to the lower bound $\lambda_{ij}^\modality{l}$ of \ac{sch}, defined in \Cref{eq:sch:lambdal}. One design goal for \ac{dsch} was continuity, which is why we opted to interpolate between the two cases in \Cref{eq:sch:lambdal} based on the similarity. This resulted in the additional term $\mathbf{S}_{ij} \cdot \tau$ in \Cref{eq:dsch:it:p}.

Finally, we combine the dynamic semantic channel width from \Cref{eq:dsch:it:w} with the updated computation of the leftmost semantic channel point from \Cref{eq:dsch:it:p} into our proposed loss presented in \Cref{eq:dsch:it:l}. Analogously to \ac{sch} (see~\Cref{eq:sch:loss}), the objective is defined as a sum over all samples and all combinations of modalities in order for the model to learn a Hamming space projection considering both intra-modal and cross-modal similarities. Furthermore, we also chose to introduce the option to enable custom weighting of the similarity sets using hyperparameters $\alpha$ and $\beta$, which is disabled by default, i.e., $\alpha = \beta = 1$, resulting in $\psi_{ij} = 1$ in all cases. Another hyperparameter is $\gamma_\ell$, which acts as a curve modifier to penalize large distance-errors, e.g., quadratically. It defaults to \num{1} for linear loss scaling. As stated previously in \Cref{s:loss:channels}, the label cosine similarity determines the Hamming distance of two hash code vectors following \Cref{eq:dis}.

\subsubsection{Vectorized Formulation} The proposed loss can be vectorized in a straight-forward manner. Following the notation in~\cite{Hu2024}, we construct matrices from elements, e.g., the matrix $\mathbf{\Psi} = \left[\psi_{ij}\right]^{n \times n}$ is constructed from elements $\psi_{ij}$, where $i$ is the row index and $j$ is the column index. We denote an appropriately sized matrix with elements all ones as $\mathbf{1}$, the Haramad product as $\odot$, and the element-wise power function as~$\circ$. The $\relu$ activation function operates on elements, too, and the $\text{L}_{1,1}$-norm is denoted by $\lVert \cdot \rVert_1$.
\begin{gather}
\begin{aligned}
    \mathcal{L}_\text{D} = \kern1.8em&\kern-1.8em \sum_{m_i, m_j}^{\{I, T\}} \left(\begin{aligned}
        &\lVert \left(\mathbf{\Psi} \odot \relu\left(\mathbf{P} - \mathbf{D}^* \right) \right)^{\circ \gamma_\ell}\rVert_1  +\\
        &\lVert \left(\mathbf{\Psi} \odot \relu\left(\mathbf{D}^* - \left(\mathbf{P} + \mathbf{W}\right) \right) \right)^{\circ \gamma_\ell}\rVert_1
    \end{aligned}\right),  \\[5pt]
    \text{where~} \mathbf{D}^* &= \left[d\left(\mathbf{\hat{B}}_i^\mmodality{m_i}, \mathbf{\hat{B}}_j^\mmodality{m_j}\right)\right]^{n \times n} \begin{array}{l}\text{the observed}\\\text{Hamming distances,}\end{array} \\[5pt]
    \text{and~} \mathbf{W} &= \left(k - \lambda_\text{neg} - \tau\right) \cdot \left(\mathbf{1} - \mathbf{S}\right)^{\circ \gamma_w} + \tau \cdot \mathbf{1}, \\
    \text{and~} \mathbf{P} &= \lambda_\text{neg} \cdot \left(\mathbf{1} - \mathbf{S}\right) - \tau \cdot \mathbf{S}. \label{eq:dsch:vec}
\end{aligned}\raisetag{\baselineskip}
\end{gather}
All operations in \Cref{eq:dsch:vec} can be implemented efficiently using matrices. We provide a reference implementation that ships with both the iterative and vectorized versions of our loss algorithm\rlap{.}\footnote{\url{https://github.com/Hirnmoder/dynamic-semantic-channel-hashing}}

\subsubsection{Additional Cross-Modal Quantization Loss} A detailed description of the experimental models is provided in \Cref{s:exp:model}. In order to allow for different final activation functions, we introduce an additional quantization $\mathcal{L}_q$ as regularization term to the \ac{dsch} objective. Inspired by \citeauthor{Zhou2023}~\cite{Zhou2023}, we define the quantization error as the distance of the model output to the cross-modal quantized target hash code. However, we employ the $\text{L}_{1,1}$-norm over the error instead of the squared Frobenius norm. We present the quantization error, which is summed over each modality, in vectorized form:
\begin{gather}
    \mathcal{L}_q = \sum_{m}^{\{I, T\}} \lVert \mathbf{\hat{B}}^\mmodality{m} - \mathbf{B} \rVert_1, \quad \text{where~} \mathbf{B} = \sign\left( \sum_{m}^{\{I, T\}} \mathbf{\hat{B}}^\mmodality{m} \right).\label{eq:dsch:quant}\raisetag{-0.5ex}
\end{gather}
This formulation of quantization error has the benefit of an additional cross-modal error term introduced by the summation of model outputs before quantization into $\mathbf{B}$. This incentivizes the model further to align the Hamming space projection of sample pairs, thus, enforcing a shared Hamming space.

\subsubsection{Overall Loss Function} In a final step, all previously defined loss terms are combined into an overall loss function. For improved stability during training, we opt to normalize the losses by batch size. Furthermore, the quantization error gets weighted by a factor of $\kappa_q = 0.01$ as it would hurt performance if weighted too strong. Our proposed overall loss function is defined as follows:
\begin{equation}
    \mathcal{L} = \frac{1}{n^2} \mathcal{L}_\text{D} + \kappa_q \cdot \frac{1}{n} \mathcal{L}_q. \label{eq:dsch:full}
\end{equation}

As referenced earlier, all used hyperparameters and their initial values are listed in \Cref{t:hyperparams}.\vspace{-0.5ex}
\begin{table}[t]
    \centering
    \caption{Hyperparameters and their initial value.}\label{t:hyperparams}
    \begin{tabular}{@{}r@{~}l@{\hspace{5pt}}p{\dimexpr\linewidth-29mm}@{}}\toprule
         \multicolumn{2}{@{}l@{~}}{\textbf{Hyperparameter}} & \textbf{Description}\\\midrule
         $\lambda_\text{neg}$ & $= \frac{k}{2}$ & Minimum distance of negative pairs \\
         $\tau$ & $= 1$ & Minimum channel width \\
         $\gamma_w$ & $= 8$ & Channel width curve modifier \\
         $\gamma_\ell$ & $= 1$ & Loss curve modifier \\
         $\alpha$ & $= 1$ & Fully positive sample pairs set weight \\
         $\beta$ & $= 1$ & Negative sample pairs set weight \\
         $\kappa_q$ & $=0.01$ & Quantization error weight\\\bottomrule
    \end{tabular}\vspace{-1ex}
\end{table}

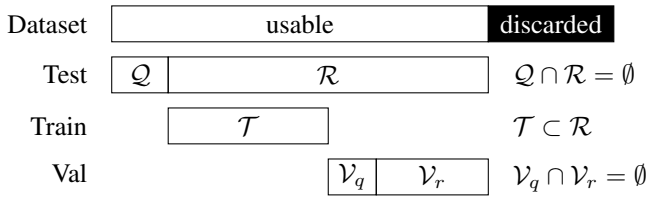
\begin{figure}[b]
    \centering\vspace{-1.5ex}
    \begin{tikzpicture}
    \tikzset{
        dataset/.style={draw, inner sep=0pt, outer sep=0pt, minimum width=#1*\linewidth, minimum height=3ex},
        invert/.style={fill=., text=.!0!},
        width/.style={execute at begin scope={\setlength\linewidth{\dimexpr#1\linewidth}}}
    }
    \begin{scope}[width=0.75]
        \node (usable) [dataset=0.75] {usable};
        \node (unusable) [right=0 of usable] [dataset=0.25, invert] {discarded};
        \begin{scope}[width=0.75]
            \node (test-q) [below right=0.2 and 0 of usable.south west] [dataset=0.15] {$\mathcal{Q}$};
            \node (test-r) [right=0 of test-q] [dataset=0.85] {$\mathcal{R}$};
            \begin{scope}[width=0.85]
                \node (train) [below right=0.2 and 0 of test-r.south west] [dataset=0.5] {$\mathcal{T}$};
                \node (val-q) [below right=0.2 and 0 of train.south east] [dataset=0.15] {$\mathcal{V}_q$};
                \node (val-r) [right=0 of val-q] [dataset=0.35] {$\mathcal{V}_r$};
            \end{scope}
        \end{scope}
    \end{scope}
    \node (lbl-dataset) [base left=0.2 of usable.base-|usable.west] {Dataset};
    \node (lbl-test) [base left=0.2 of test-q.base-|usable.west] {Test};
    \node (lbl-train) [base left=0.2 of train.base-|usable.west] {Train};
    \node (lbl-val) [base left=0.2 of val-q.base-|usable.west] {Val};
    \node (set-test) [base right=0.2 of test-q.base-|usable.east] {$\mathcal{Q} \cap \mathcal{R} = \emptyset$};
    \node (set-train) [base right=0.2 of train.base-|usable.east] {$\mathcal{T} \subset \mathcal{R}$};
    \node (set-val-r) [base right=0.2 of val-r.base-|usable.east] {$\mathcal{V}_q \cap \mathcal{V}_r = \emptyset$};
\end{tikzpicture}%
    \caption{Schematic representation of the dataset partitioning process to obtain query, retrieval, train, validation query, and validation retrieval sets.}\label{f:data-partitions}
\end{figure}

\section{Experiment Setup}\label{s:exp}
\subsection{Data Material and Data Partitioning}\label{s:exp:data}
We perform experiments on two well-known cross-modal multi-label benchmark datasets, namely \NUSWIDE~\cite{Chua2009} and \MIRFLICKR~\cite{Huiskes2008}. Both provide samples comprising an image, its user-provided hashtag list, and a set of manually annotated category labels.

\begin{table}[b]
    \centering
    \caption{Number of samples per dataset and data partition.}\label{t:data-partitions}
    \sisetup{
        table-alignment-mode=format,
        table-format=6.0,
        table-number-alignment=center,
        table-text-alignment=center,
    }
    \begin{tabular}{r@{~}>{$}l<{$}SS}\toprule
        \multicolumn{2}{r!{~}}{\multirow[c]{2}{*}[-2pt]{\bfseries Partition}} & \multicolumn{2}{c}{\bfseries Numbers of samples} \\\cmidrule{3-4} 
                             &                                                & {\NUSWIDE} & {\MIRFLICKR} \\\midrule
        Total                &                                                & 269648     & 25000 \\
        Usable               & \mathcal{D}                                    & 195834     & 20015 \\
        (Test) Query         & \mathcal{Q}                                    & 2100       & 2000 \\
        (Test) Retrieval     & \mathcal{R}                                    & 193734     & 18015 \\
        Train                & \mathcal{T}                                    & 10500      & 10000 \\
        Validation Query     & \mathcal{V}_q                                  & 2100       & 2000 \\
        Validation Retrieval & \mathcal{V}_r                                  & 181134     & 6015 \\\bottomrule
    \end{tabular}
\end{table}

\NUSWIDE~\cite{Chua2009} contains \num{269648} samples, each consisting of one image and \numrange{0}{632} hashtags. Additionally, each sample is annotated with \numrange{0}{13} out of a set of \num{81} labels. In accordance to some previous research~\cite{Zhou2023,Wang2024,Jiang2017,Luo2023}, we select the \mbox{top-\num{21}} labels and discard any samples with no labels, resulting in a subset of \num{195834} samples. It is worth noting that in the literature there is no consensus about the subset selection process and some select the \mbox{top-\num{10}} labels~\cite{Li2024,Zhou2023a,Hu2024} while others seem to use the full dataset~\cite{Hong2022}. We then randomly sample \num{2100} samples to form the query set~$\mathcal{Q}$ with the remaining samples forming the retrieval set~$\mathcal{R}$. Next, a total of \num{10500} samples are randomly selected from $\mathcal{R}$ to form the training set~$\mathcal{T}$. Mathematically, $\mathcal{T} \subset \mathcal{R}$ and $\mathcal{R} \cap \mathcal{Q} = \emptyset$. Although literature suggests sampling \num{100} instances per label as queries and \num{500} instances per label for training~\cite{Luo2023} or \enquote{an equal number of samples from all [labels],}~\cite{Hu2024} this is an unsatisfactory solution to balance label occurrence probabilities. On the contrary, this naïve approach leads to an even more unbalanced dataset due to multi-label interdependence. We, therefore, opted to use random sampling with fixed seed to closely match the dataset distribution instead of introducing skew or using other methods, such as iterative stratification~\cite{Sechidis2011}.

\MIRFLICKR~\cite{Huiskes2008} contains \num{25000} samples comprising one image and \numrange{0}{75} hashtags as well as \numrange{0}{14} labels out of a set of \num{24} labels. We follow a common subset selection process~\cite{Zhou2023,Wang2024,Jiang2017,Hu2024} and select all samples that have at least one hashtag that is included in the \mbox{top-\num{20}} hashtags. Furthermore, all samples that have no labels are discarded, resulting in a subset of \num{20015} samples. Finally, we employ the same subsampling strategy as with \NUSWIDE and select \num{2000} samples for $\mathcal{Q}$ and \num{10000} samples for $\mathcal{T}$ in order to comply with other implementations~\cite{Zhou2023,Hu2024,Jiang2017}.

For model selection, hyperparameter tuning, and to assess model performance during development, we employ two additional partitions: the validation query set $\mathcal{V}_q$ and the validation retrieval set $\mathcal{V}_r$. To the best of our knowledge, there is no evidence in the literature for performing such train-validation-test split in the cross-modal retrieval domain. Our train-test-validation splitting approach is designed to accomplish \numberstringnum{\getrefnumber{e:data:last}} major goals:
\begin{enumerate*}
    \item perform validation on a query set with a similar size as the (test) query set,
    \item make validation and train sets completely disjoint to assess model generalization capabilities, and
    \item\label{e:data:last} keep test and train set construction identical to existing literature for better comparison.
\end{enumerate*}
To construct the validation partition, we first take the remainder of the retrieval set without training samples $\mathcal{R} \setminus \mathcal{T}$. Next, depending on the initial dataset being \NUSWIDE or \MIRFLICKR, a total of \num{2100} or \num{2000} samples are selected at random to form the validation query set $\mathcal{V}_q$, respectively. The remaining samples are then assigned to the validation retrieval set, i.e., $\mathcal{V}_r = \mathcal{R} \setminus \mathcal{T} \setminus \mathcal{V}_q$. The breakdown of the total dataset into all data partitions is depicted in \Cref{f:data-partitions}. Furthermore, the number of samples for each data partition is reported in \Cref{t:data-partitions}.

\subsection{Model Architecture and Data Preprocessing}\label{s:exp:model}
To demonstrate the effectiveness of our proposed loss function, we implement two model architectures, both of which use the Transformer architecture~\cite{Vaswani2017}. In the following, both architectures are outlined and their respective data preprocessing steps are described.

As a baseline, we employ a dual-stream model proposed by \citeauthor{Zhou2023}~\cite{Zhou2023} named \TDSRDH. Its image stream consists of a pre-trained ResNet-152 backbone and a three-layer hash-learning \ac{mlp}. The text stream consists of a single-layer Transformer with four heads followed by a three-layer hash-learning \ac{mlp}. The authors introduce a three-fold loss comprising a cross-modal pairwise loss, a quantization loss, and a cross-modal triplet loss~\cite{Zhou2023}. We implemented their architecture and loss function based on the descriptions in their paper as no code is publicly available. This is also true for data preprocessing. Furthermore, we were unable to obtain a \num{512}-dimensional \model{Word2Vec}~\cite{Mikolov2013} word embedding model mentioned by~\citeauthor{Zhou2023}~\cite{Zhou2023}. Instead, we chose the \model{Google News 300d}\footnote{\url{https://code.google.com/archive/p/word2vec/}} embeddings and added a trainable linear layer to upscale the \num{300}-dimensional word embeddings to the \num{512} dimensions used in the Transformer. Next, we apply sinusoidal positional encoding as proposed by \citeauthor{Vaswani2017}~\cite{Vaswani2017}. However, as described in \Cref{s:exp:data}, the datasets' texts are essentially user-provided hashtags and not ordinary descriptions. In order to cleanse and tokenize these texts as well as correct misspellings, we use \software{symspellpy}~\cite{mammothb2025}. The maximum sequence length is configurable and set to \num{128} tokens.

The second model architecture follows a hybrid-stream approach. As the multi-modal backbone we employ an off-the-shelf CLIP model~\cite{Radford2021}, namely \model{ViT-H-14-quickgelu}, pre-trained on the \dataset{DFN-5B} dataset~\cite{Fang2023} and implemented by OpenCLIP~\cite{Ilharco2021}. We then append a \ac{mlp} to map the CLIP embeddings to the output Hamming space. We call this model \CLIPHash. To the best of our knowledge, this architectural design has not previously been explored in the context of image-text Semantic Hashing. A similar model architecture was proposed for the video-text domain~\cite{Zhuo2022}, and proposals to the image-text domain either use \ac{bow} text inputs~\cite{Mingyong2023}, a different text-stream model~\cite{Li2023a,Li2024}, labels instead of descriptions as text input~\cite{Cao2024}, or cannot be used in a cross-modal retrieval setting~\cite{Xia2023}. Since OpenCLIP provides their own text preprocessing pipeline including tokenization and embedding, no custom preprocessing steps are implemented for \CLIPHash. The maximum sequence length for the selected CLIP backbone is \num{77} tokens.

\begin{figure*}[htbp]
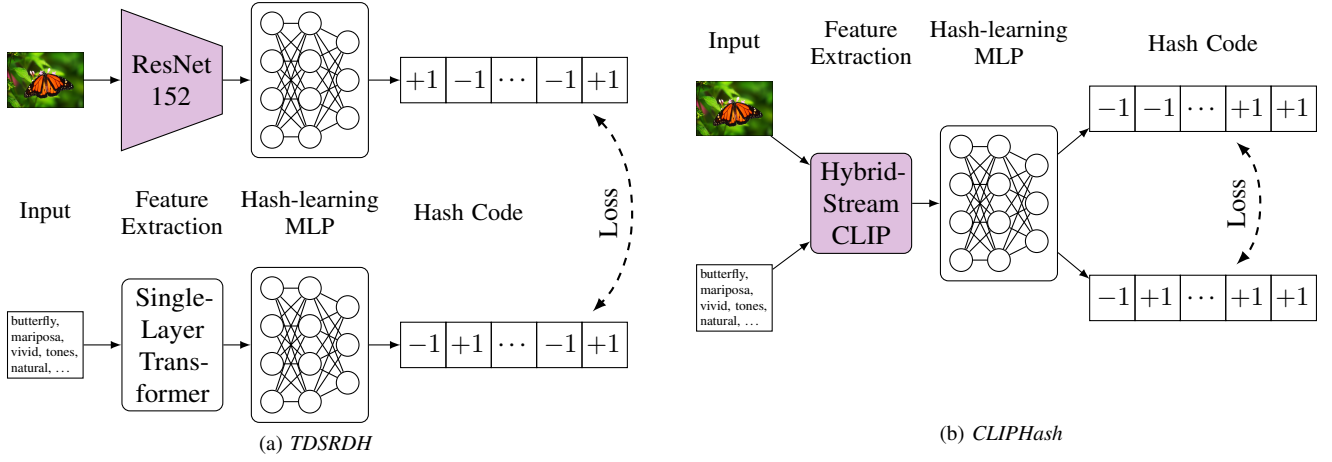

\centering
\subfloat[\TDSRDH]{\begin{minipage}{0.5\linewidth}%
    \centering%
    \input{model-tdsrdh}%
    \label{f:modelarch:tdsrdh}%
\end{minipage}}%
\subfloat[\CLIPHash]{\begin{minipage}{0.5\linewidth}%
    \centering%
    \input{model-cliphash}%
    \label{f:modelarch:cliphash}%
\end{minipage}}
\caption{Schematic overview over the two model architectures employed for the experiments in this paper. Shaded in purple are off-the-shelf pre-trained modules, which are frozen during training.}
\label{f:modelarch}
\end{figure*}

In \Cref{f:modelarch}, the two model architectures as well as an example forward flow are depicted. In both \namecrefs{f:modelarch}, purple shaded parts mark frozen, off-the-shelf pre-trained modules in each architecture. The dual-stream structure of \TDSRDH is clearly visible in \Cref{f:modelarch:tdsrdh} and the loss function acts as the only \enquote{connecting tissue} between modalities. This is different from the hybrid-stream structure of \CLIPHash depicted in \Cref{f:modelarch:cliphash}. Here, both modalities share some layers within the pre-trained CLIP model and -- more importantly -- they share a common hash-learning \ac{mlp}. Furthermore, the loss function imposes additional cross-modal alignment constraints.

The configuration of the hash-learning \acp{mlp}, that is their number of layers and the neurons per layer, is dependent on the model and experiment. However, the defaults for \TDSRDH are the configurations given in \cite{Zhou2023}. For example, we denote a three-layer \ac{mlp} with \num{512}-dimensional inputs, a first hidden layer with \num{1024} neurons, a second hidden layer with \num{8192} neurons, and an output of dimension $k$ as follows: \mbox{\mlpcfg[\tanh/\text{d}=\num{0.1},\tanh/\text{d}=\num{0.1}]{512,1024,8192,k}}. We also specify the activation functions and dropout probabilities used after each layer. In this example, the identity function, i.e., no activation function, is used for the last layer. Contrary to \citeauthor{Zhou2023}~\cite{Zhou2023} and \citeauthor{Hu2024}~\cite{Hu2024}, we employ the $\tanh$ activation function in the inner layers of the hash-learning \ac{mlp} and use no output activation function in order for the model to be able to express semantic subtleties with outputs exceeding the \num{\pm 1} bound that would be imposed by the $\tanh$ function. Dropout is never applied to the last layer.

The image-stream hash-learning \ac{mlp} of \TDSRDH has the following configuration based on \cite{Zhou2023}, where \num{2048} is the output size of ResNet-152: \mbox{\mlpcfg[\tanh/\text{d}=\num{0.1},\tanh/\text{d}=\num{0.1}]{2048,4096,4096,k}}. Similarly, following \cite{Zhou2023} the text-stream hash-learning \ac{mlp} is configured as follows: \mbox{\mlpcfg[\tanh/\text{d}=\num{0.1},\tanh/\text{d}=\num{0.1}]{512,1024,8192,k}}.

The \CLIPHash model contains only a single hash-learning \ac{mlp}, which retrieves a \num{4096}-dimensional embedding vector from the backbone and processes it in five layers as follows: \mbox{\mlpcfg[\tanh/\text{d}=\num{0.1},\tanh/\text{d}=\num{0.1},\tanh/\text{d}=\num{0.1},\tanh/\text{d}=\num{0.1}]{4096,4096,4096,1024,256,k}}.

\subsection{Model Training}\label{s:exp:train}
The training procedure is implemented as a straight-forward batched learning strategy to optimize the model parameters using gradient descent. Although the \ac{dsh} literature predominately reports the use of the SGD optimizer (see~\cite{Hu2024,Zhou2023,Li2023a,Deng2018,Mingyong2023}), we utilize the Adam optimizer~\cite{Kingma2017} for faster convergence as do some recent papers (see~\cite{Huang2024,Li2023,Qiu2018}). We configure Adam with an initial learning rate of \num{1e-5}, an epsilon set to \num{1e-8}, no weight decay, and momentum hyperparameters set to $\beta_1=0.9$ and $\beta_2=0.999$. We train for \num{200} epochs and use a custom learning rate scheduler that applies a cosine-annealing drop-off from \numrange{1e-5}{1e-6} between epochs \numrange{75}{150}. The batch size is set to \num{128}.

During training, we perform data augmentations to the image modality to enhance the diversity and variance of the training dataset as well as to incentivize the model to generalize on unseen data. We define four hardness levels of data augmentations and employ a progressive augmentation mixture strategy with fixed ratios at each step, which are listed in \Cref{t:progressiveaug}. The training starts with no augmentation at all and continues to add a growing portion of augmentations with increasing hardness, while also retaining \qty{25}{\percent} unaltered images. The application of data augmentation in the \ac{dsh} domain is uncommon since only \citeauthor{Luo2023}~\cite{Luo2023} report a few training procedures with image augmentation, whereas the other surveys \cite{Wang2024,Singh2022,Zhou2023a} contain no notice of it. However, augmenting training samples is standard practice in the Computer Vision domain~\cite{Xu2023a}, which is why we opted to implement model-free single-image as well as model-free non-instance-level multi-image augmentations. For each hardness level, we define one or more groups of augmentations as well as a count parameter~\# that indicates how many individual augmentations per group can be active. All augmentation groups and their count parameter as well as the individual augmentations and their configurations are listed in \Cref{t:augmentations} clustered by hardness level.

\begin{table}[!b]%
    \vspace{-15ex}
    \begin{minipage}{\linewidth}
        \centering
        \caption{Progressive image augmentation mixtures.}\label{t:progressiveaug}
        \begin{tabular}{ccccc}\toprule
            \multirow{2}{1.3cm}[-2pt]{\centering\textbf{Epoch Range}} & \multicolumn{4}{c}{\textbf{Image Augmentation Mixture}} \\\cmidrule{2-5} 
                                        & None                & Easy               & Medium             & Hard \\\midrule
            \phantom{0}\numrange{0}{19} & \qty{100}{\percent} & \qty{0}{\percent}  & \qty{0}{\percent}  & \qty{0}{\percent} \\
            \numrange{20}{39}           & \qty{25}{\percent}  & \qty{75}{\percent} & \qty{0}{\percent}  & \qty{0}{\percent} \\
            \numrange{40}{59}           & \qty{25}{\percent}  & \qty{50}{\percent} & \qty{25}{\percent} & \qty{0}{\percent} \\
            \numrange{60}{79}           & \qty{25}{\percent}  & \qty{25}{\percent} & \qty{50}{\percent} & \qty{0}{\percent} \\
            \numrange{80}{99}           & \qty{25}{\percent}  & \qty{25}{\percent} & \qty{25}{\percent} & \qty{25}{\percent} \\
            \numrange{100}{149}         & \qty{25}{\percent}  & \qty{0}{\percent}  & \qty{25}{\percent} & \qty{50}{\percent} \\
            \numrange{150}{199}         & \qty{25}{\percent}  & \qty{0}{\percent}  & \qty{0}{\percent}  & \qty{75}{\percent} \\\bottomrule
        \end{tabular}
    \end{minipage}\par\vspace{3ex}
    \begin{minipage}{\linewidth}
        \centering
        \caption{Image augmentation groups by hardness level.}\label{t:augmentations}%
        \begin{tabular}{@{}c@{~}cp{29mm}@{~}p{44mm}@{}}\toprule
            \multirow{2}*[-2pt]{\textbf{Level}} & \multirow{2}*[-2pt]{\textbf{\#}} & \multicolumn{2}{c}{\textbf{Augmentations}}\\\cmidrule(l){3-4}
            && Description & Configuration \\\midrule

            None & \num{0} & --- & --- \\\midrule

            \multirow{4}*{\rotatebox{90}{Easy}} & \multirow{4}*{\num{1}} & Flip horizontally & $p=1$ \\
            && Rotate by fixed value & \ang{\pm90}, \ang{\pm45}\\
            && Convert to grayscale & --- \\
            && Random crop & $\text{scale} \in [0.5, 1.0], \text{ratio}=1$ \\\midrule

            \multirow{7}*[-3pt]{\rotatebox{90}{Medium}} & \num{1} & Flip horizontally & $p = 0.5$ \\\cmidrule{2-4}
            & \multirow{6}*{\rotatebox{90}{\numrange{1}{2}}} & Random rotation between & \qtyrange{-45}{45}{\degree} \\
            && Convert to grayscale & --- \\
            && Random crop & $\text{scale} \in [0.35, 1.1], \text{ratio} \in [0.9, 1.1]$ \\
            && Random erasing & $\text{scale} \in [0.02, 0.05], \text{ratio} \in \left[\frac{1}{3}, 3\right]$ \\
            && Random color jitter\textsuperscript{$*$} & $\text{b, c, s} \in [0.8, 1.2], \text{h} \in [-0.1, 0.1]$ \\
            && Random uniform noise & $\text{blending} \in [0.0, 0.5]$\\\midrule

            \multirow{11}*[-10pt]{\rotatebox{90}{Hard}} & \num{1} & Flip horizontally & $p = 0.5$ \\\cmidrule{2-4}
            & \multirow{8}*{\rotatebox{90}{\numrange{1}{3}\textsuperscript{$\dagger$}}} & Random rotation between & \qtyrange{-90}{90}{\degree}\\
            && Convert to grayscale & --- \\
            && Random crop & $\text{scale} \in [0.2, 1.1], \text{ratio} \in \left[\frac{3}{4}, \frac{4}{3}\right]$ \\
            && Random erasing & $\text{scale} \in [0.05, 0.15], \text{ratio} \in \left[\frac{1}{4}, 4\right]$ \\
            && Random color jitter\textsuperscript{$*$} & $\text{b, c, s} \in [0.5, 1.5], \text{h} \in [-0.25, 0.25]$ \\
            && Flip vertically & $p=1$ \\
            && Elastic transform & $\alpha=50, \sigma=5$ \\
            && Random uniform noise & $\text{blending} \in [0.0, 0.9]$\\\cmidrule{2-4}
            & \multirow{2}*{\num{1}\textsuperscript{$\dagger$}} & Full-frame image blend & $\text{blending} \in [0.0, 0.45]$\\
            && Picture-in-picture & $\text{scale} \in [0.05, 0.25]$ \\\bottomrule\addlinespace[\belowrulesep]

            \multicolumn{4}{@{\hspace{4pt}}p{\dimexpr\linewidth-10pt}}{\# is the count parameter, indicating how many augmentations of this group\newline\phantom{\#} are active at any one time.} \\
            \multicolumn{4}{@{\hspace{4pt}}l}{\textsuperscript{$*$} $\text{b, c, s, h}$ are brightness, contrast, saturation, and hue, respectively.} \\
            \multicolumn{4}{@{\hspace{4pt}}p{\dimexpr\linewidth-10pt}}{\textsuperscript{$\dagger$} Only one of the two augmentation groups is active per sample. The upper\newline\phantom{\textsuperscript{$\dagger$}} group has a probability of \qty{80}{\percent} to be selected, the lower group has \qty{20}{\percent}.}
        \end{tabular}
        \vspace*{-1ex}
    \end{minipage}
\end{table}

In order to demonstrate the efficacy of our proposed loss function, we also train the models with different losses. For fair comparison we try to match the training procedure and hyperparameter setup of the original papers. As mentioned in \Cref{s:exp:model}, the authors of the \TDSRDH model proposed a three-fold loss function alongside~\cite{Zhou2023}. When training using their TDSRDH-loss, we also follow their alternate-learning strategy~\cite{Zhou2023}, which means that two streams -- image and text -- get updated in an alternating manner. However, we use the Adam optimizer as stated above, and introduce a normalization to the loss terms, similar to the ones of our proposed loss described in \Cref{s:loss:proposed} and, in particular, \Cref{eq:dsch:full}. This also requires a re-evaluation of the TDSRDH-loss weighting hyperparameters, which we found matching the original paper's performance claim when set to \mbox{$\alpha_\text{\TDSRDH}=\num{0.05}$} and \mbox{$\beta_\text{\TDSRDH}=\num{1}$}. The batch size, initial learning rate, and number of epochs match our settings described above.

When training with the \ac{sch} loss by \citeauthor{Hu2024}~\cite{Hu2024}, we use their hyperparameter setup and set $\alpha_\text{\ac{sch}} = \num{1}, \beta_\text{\ac{sch}} = \num{1}$ and $\tau_\text{\ac{sch}} = \num{3}$. The training routine is identical to ours and updates both image and text stream simultaneously. In the accompanying code, the authors also employ normalization to the loss terms~\cite{Hu2024}, which is similar to ours. However, we use the Adam optimizer with a different learning rate, a larger batch size of \num{128} instead of \num{32}, and no weight decay in order to achieve better comparability with our proposed loss.

It is important to note that we conduct each experiment at least three times with different initializations to obtain an averaged result for a decreased effect of model training randomness. This also allows us to assess model performance spread induced by random initializations.

\subsection{Evaluation Metrics}\label{s:exp:eval}
For assessing the cross-modal retrieval performance of \ac{dsh} models, the literature has agreed on several metrics with \ac{map} and the precision-recall curve being one of the more popular ones~\cite{Wang2024,Luo2023,Singh2022}. Both metrics are also reported by the authors of the \TDSRDH model~\cite{Zhou2023} and the \ac{sch} loss~\cite{Hu2024}. Although these two metrics are used throughout the literature, there exist many different interpretations as to how to compute them and what specifications to report. This is especially true for \ac{map}, where the cutoff (often reported as top-$k$ or \acs{map}@$k$~\cite{Wang2024}) differs hugely across papers -- ranging from~\numrange[parse-numbers=false]{50}{5000} or no cutoff -- see for example~\cite{Hu2024,Jiang2017,Deng2018,Hong2022,Li2023a,Mingyong2023,Huang2024,Li2024,Qiu2018}. These inconsistencies across the literature effectively void model performance comparisons as different cutoffs result in different scores. \citeauthor{He2018}~\cite{He2018} report an overestimation even when using a high cutoff value of \num{5000} compared to the true \ac{map}. Furthermore, the authors point out a significant issue with the \ac{map} formulation found throughout the literature: it is not permutation-agnostic~\cite{He2018,McSherry2008}. To clarify the issue, we take a look at the formula for calculating \ac{map}, which is given in the literature~\cite{Wang2024,Singh2022,Luo2023}. To avoid confusion with the hash code length $k$, here the cutoff value is denoted as $z$ with $1 \leq z \leq \lvert \mathcal{R}\rvert$:
\begin{equation}
\begin{aligned}
    \text{mAP}@z &= \frac{1}{\lvert \mathcal{Q}\rvert}\sum_{j=1}^{\lvert \mathcal{Q}\rvert} \text{AP}@z(q_j), \\
    \text{where~} z & > 0 \text{~the cutoff value}\\
    \text{or~} z & =\lvert\mathcal{R}\rvert \text{~without cutoff}, \\
    \text{and~} q_j & \in \mathcal{Q} \text{~elements of the query set.} \label{eq:lit:map}
\end{aligned}
\end{equation}
The \ac{map} is the samples' \ac{ap} value at the specified cutoff, averaged over all samples in the query set. The \ac{ap} for one sample $q \in \mathcal{Q}$ can be calculated as follows:\vspace{2pt}
\begin{gather}
\begin{aligned}
    \text{AP}@z(q)  &= \frac{\sum_{i=1}^{z} \text{P}@i(q)\cdot \delta(q, r_i)}{\sum_{i=1}^{z} \delta(q, r_i)}, \\[3pt]
    \text{where~} \text{P}@i(q) &= \frac{1}{i}\sum_{t=1}^{i} \delta(q, r_t) \text{~the top-}i \text{~retrieval precision},\\
    \text{and~} \delta(q, r) &= \begin{cases*}
        1 & if samples $q$ and $r$ are label-similar, \\
        0 & otherwise,
    \end{cases*}\\
    \text{and~} r_i & \in \mathcal{R} \text{~the~} i^\text{th}\text{~retrieved sample.} \label{eq:lit:ap}
\end{aligned}\raisetag{\baselineskip}
\end{gather}
The $\delta$ function can be defined more formal for two samples $x_i$ and $x_j$ with labels $l_i$ and $l_j$ as follows:
\begin{equation}
    \delta(x_i, x_j) = \begin{cases*}
        1 & if $\mathbf{S}_{ij} > 0$,\\
        0 & if $\mathbf{S}_{ij} = 0$,
    \end{cases*} = \begin{cases*}
        1 & if $l_i \cdot l_j > 0$, \\
        0 & if $l_i \cdot l_j = 0$.
    \end{cases*} \label{eq:lit:delta}
\end{equation}
The issue with this definition of \ac{ap} lies in the ordering of the retrieval samples $r_i$. In the context of \ac{dsh}, each sample gets assigned a discrete hash code within the Hamming space, thus, distances between samples are also discrete:
$$d\left(b_i^\mmodality{m_i}, b_j^\mmodality{m_j}\right) \in \{0, \dots, k\}.$$
Since the set of possible distances has exactly $k+1$ elements, but in practice $\lvert \mathcal{R}\rvert > k+1$, there must exist at least two samples $r_i, r_j$ with $i \ne j$ having the same Hamming distance from a given query sample. Then, there is no semantic ordering among $r_i$ and $r_j$, thus, it is undefined which of the two samples appears first in the retrieval process. This ambiguity makes the common formulation of \ac{ap} susceptible to randomness and poses the possibility of maliciously adjusting the retrieval sequence to achieve higher scores. To demonstrate this, let's consider the following three semantically-equivalent retrieval sequences for a single query depicted in \Cref{f:map:retrievalsequences}:

\begin{figure}[!h]
\centering\def\positivesign{$\oplus$}\def\negativesign{$\ominus$}\vspace{-1.5ex}%
\begin{tikzpicture}
    \foreach \d/\e/\f [count=\y from 1] in {\P\P\N/\P\N\N/\P\N, \N\P\P/\N\N\P/\N\P, \P\N\P/\N\P\N/\P\N} {
        \coordinate (s\y-0) at (0, -0.7*\y);
        \foreach \a [count=\x from 1, remember=\x as \xl initially 0] in {\d, \e, \f} {
            \edef\P{\;\positivesign\;}\edef\N{\;\negativesign\;}
            \node (s\y-\x) [right=2pt of s\y-\xl] {\a~};
        }
        \node [fit=(s\y-1)(s\y-3), inner ysep=1pt, inner xsep=3pt, draw] {};
    }
    \tikzset{
        dgrp/.style={inner xsep=-1pt, inner ysep=5pt, yshift=2pt, draw, dashed, rounded corners=3pt, outer sep=0pt}
    }
    \path node (lbl-d0) [above=0.1 of s1-1] {$d=0$}
          node (lbl-d1) [above=0.1 of s1-2] {$d=1$}
          node (lbl-d2) [above=0.1 of s1-3] {$d=2$}
          node (bor-d0) [fit=(s1-1)(s3-1)(lbl-d0.center), dgrp] {}
          node (bor-d1) [fit=(s1-2)(s3-2)(lbl-d1.center), dgrp] {}
          node (bor-d2) [fit=(s1-3)(s3-3)(lbl-d2.center), dgrp] {}
          node (lbl-s1) [left=0.15 of s1-1] {Seq.\,1:}
          node (lbl-s2) [left=0.15 of s2-1] {Seq.\,2:}
          node (lbl-s3) [left=0.15 of s3-1] {Seq.\,3:}
          node (ap-s1) [right=0.15 of s1-3] {$\text{AP} = \num{0.830}$}
          node (ap-s2) [right=0.15 of s2-3] {$\text{AP} = \num{0.542}$}
          node (ap-s3) [right=0.15 of s3-3] {$\text{AP} = \num{0.710}$}
    ;
\end{tikzpicture}\vspace{-1.5ex}%
\caption{Demonstration of different retrieval sequences that are semantically equivalent, but result in significantly different \acl{ap} values. Retrieval samples that are similar to the query are represented as~\positivesign, and dissimilar retrieval samples as~\negativesign. The example is inspired by \citeauthor{He2018}~\cite{He2018}.}\label{f:map:retrievalsequences}\vspace{-1ex}%
\end{figure}

All three sequences are equivalent from a semantics standpoint, i.e., the sets defined by the Hamming distance to the query sample contain the same retrieval samples and are, therefore, identical. The only difference between the sequences is their internal ordering of samples within each Hamming distance set. Nevertheless, when calculating the \ac{ap} using \Cref{eq:lit:ap}, the results ranging from \numrange{0.542}{0.830} are vastly different. To address this issue, \citeauthor{He2018}~\cite{He2018} proposed a tie-aware calculation of the \ac{map} metric for the field of \ac{dsh}, which is the application of the formulations of \citeauthor{McSherry2008}~\cite{McSherry2008}. Based on both~\cite{He2018} and~\cite{McSherry2008}, we define the tie-aware \acl{ap} as $\text{AP}_\text{T}$ using our notation:
\begin{gather}
\begin{aligned}
    \text{AP}_\text{T}(q) \kern1.5em&\kern-1.5em= \frac{1}{N^+} \sum_{d=0}^{k} \begin{dcases*}
        0 & \kern-2.45em if $n_d = 0$,\\
        n_d^+ \cdot \frac{N_d^+}{N_d} & \kern-2.45em if $n_d = 1$,\\
        \frac{n_d^+}{n_d} \sum_{t=0}^{n_d - 1} \frac{1 + N_{d-1}^+ + t \cdot \frac{n_d^+ - 1}{n_d - 1}}{1 + t + N_{d-1}} & o/w,\\
    \end{dcases*}\\
    \text{where~} N^+ &= \textstyle\sum_{i=1}^{\lvert\mathcal{R}\rvert}\delta(q, r_i) \text{~the number of positive samples},\\
    \text{and~} n_d &= \lvert\{r_i \mid d(q, r_i) = d\}\rvert, \\
    \text{and~} n_d^+ &= \lvert\{r_i \mid d(q, r_i) = d \land \delta(q, r_i) = 1\}\rvert, \\
    \text{and~} N_d &= \textstyle\sum_{j=0}^d n_j \text{~the cumulative sum over~} n_d, \\
    \text{and~} N_d^+ &= \textstyle\sum_{j=0}^d n_j^+ \text{~the cumulative sum over~} n_d^+. \label{eq:tmap:ap}
\end{aligned}\raisetag{\baselineskip}
\end{gather}
In \Cref{eq:tmap:ap}, $n_d$ is the number of samples $r \in \mathcal{R}$ that have a Hamming distance of $d$ from the query $q$, and $n_d^+$ denotes the number of similar (positive) samples $r \in \mathcal{R}$ that have said Hamming distance of $d$ from the query $q$. As expected, all sequences from the example in \Cref{f:map:retrievalsequences} yield the same score of $\text{AP}_\text{T} = \num{0.691}$, which falls between the limits set by the \ac{ap} scores for Sequence~1 and Sequence~2. When using pre-shuffled datasets, the discrepancies between the tie-oblivious $\text{AP}$ and the tie-aware $\text{AP}_\text{T}$ scores is small, since a random permutation is likely to represent the average case. This is also why the reported tie-aware score is close to the tie-oblivious score of Sequence~3 in \Cref{f:map:retrievalsequences} (\num{0.691} vs. \num{0.710}).

The tie-aware formula for \acl{map} averages, analogously to \Cref{eq:lit:map}, the tie-aware \ac{ap} scores over all samples in the query set:
\begin{equation}
    \text{mAP}_\text{T} = \frac{1}{\lvert\mathcal{Q}\rvert} \sum_{j=1}^{\lvert\mathcal{Q}\rvert} \text{AP}_\text{T}(q_j). \label{eq:tmap:map}
\end{equation}

In our reference implementation, we provide the tie-aware \ac{map} score calculation in two variants. The first implementation is closely following \Cref{eq:tmap:ap,eq:tmap:map}, whereas the second one is an optimized and vectorized version partly based on the code provided by \citeauthor{He2018}~\cite{He2018}.

\section{Experiments, Results and Discussion}\label{s:res}
In the first experimental setting, models based on the \TDSRDH model architecture are trained on the proposed \ac{dsch} objective as well as two reference loss functions: \ac{sch}~\cite{Hu2024} and TDSRDH-loss~\cite{Zhou2023}. To further demonstrate the effectiveness of \ac{dsch}, experiments were conducted on both datasets and a total of four different output hash code lengths $k \in \{16, 32, 64, 128\}$. Following common notation, we denote a retrieval task with query modality~$\text{M}_q$ and retrieval modality~$\text{M}_r$ as $\text{M}_q \to \text{M}_r$.  We report the tie-aware \ac{map} scores on all four possible retrieval tasks, i.e., the two cross-modal retrieval tasks \itot\ and \ttoi\ as well as the two single-modal retrieval tasks \itoi\ and \ttot. The results of this first experimental setting on the test query and retrieval sets are reported in \Cref{t:e1:map}. They show a clear improvement in achieved scores of our proposed loss \ac{dsch} over the two reference losses in almost every scenario. Especially in the cross-modal retrieval tasks, \ac{dsch} outperforms \ac{sch} and TDSRDH-loss across all hash code lengths with, in parts, a substantial margin. In a total of three out of the \num{32} scenarios, \ac{dsch} performs subpar, however, the achieved results are only marginally worse than the respective bests. Moreover, the results of \ac{dsch} show only small deviations in many scenarios, indicating improved model stability regardless of initialization randomness compared to the competing losses. In summary, these results demonstrate the effectiveness of \ac{dsch} to guide model training towards discriminative and similarity-preserving hash codes of various lengths and across datasets that can be utilized in both cross-modal and single-modal tasks.

In order to find the optimal value for the channel width curve modifier hyperparameter $\gamma_w$, a second set of experiments was conducted. Again, the \TDSRDH model architecture and both datasets were considered. The hyperparameter search used a fixed hash code length of \mbox{$k=32$} and the scores are calculated on the validation set. We conducted experiments on a total of seven possible values for $\gamma_w$. Since the tie-aware \ac{map} scores are very similar for many hyperparameter values, no clear optimal $\gamma_w$ could be found. Indeed, \Cref{t:e2:map} shows the top-3 performances per category highlighted and setting the hyperparameter to \num{6}, \num{8}, or \num{14} seems arbitrary. Since other precision-based metrics have shown similar inconclusiveness, we opted to report the ROC-AUC-score for these experiments as well in \Cref{t:e2:ra}. There, it is evident that $\gamma_w=8$ is the optimal choice for this hyperparameter, which is the reason behind using this value in all other experiments and in \Cref{t:hyperparams}.

So far, the experiments were conducted on the dual-stream \TDSRDH model architecture. In a third experimental setting, models following the \CLIPHash model architecture were trained with all three loss functions and a fixed hash code length of $k=32$. Again, both datasets were considered and when using \ac{dsch}, we set the hyperparameter $\gamma_w = 8$. The tie-aware \ac{map} scores on the test set are reported in \Cref{t:e3:map} for both model architectures to allow for easier comparison. Similar to the first experimental setting, models trained on the proposed loss \ac{dsch} performed -- with exception to the \ttot\ task -- consistently and considerably better than models trained on the reference losses \ac{sch}~\cite{Hu2024} and TDSRDH-loss~\cite{Zhou2023}, supporting our claim of improved hash code learning using the dynamic semantic channel width approach. Furthermore, the results in \Cref{t:e3:map} clearly show that the larger \CLIPHash models are significantly more capable of generating good hash codes than the smaller \TDSRDH models regardless of training loss function. This behavior indicates the possibility of better hash code generation by further optimizing and tailoring model architectures in a future work.

\begin{table*}
    \centering
    \robustify\boldmath
\sisetup{
    detect-weight=true,
    table-format=2.2(2),
    table-model-setup=\boldmath,
    table-alignment-mode=format,
    table-number-alignment=left,
    table-text-alignment=center,
    reset-math-version=false,
}
\caption{Comparison of three loss functions by tie-aware mean average precision ($\text{mAP}_\text{T}$) scores (in percent) for four tasks on the test set. Best performance per task, dataset, and hash code length is highlighted in boldface.}
\label{t:e1:map}
\begin{tabular}{@{\hspace{3pt}}l@{\hspace{5pt}}l@{\hspace{8pt}}S*{3}{@{\hspace{5pt}}S}@{\hspace{12pt}}S*{3}{@{\hspace{5pt}}S}@{\hspace{3pt}}}
\toprule
{\multirow[b]{2}{*}{\rotatebox{90}{Task}}} & {{\hfill}Dataset\kern-8pt} & \multicolumn{4}{c}{\MIRFLICKR} & \multicolumn{4}{c}{\NUSWIDE} \\
{} & {Loss{\hfill}$k$\kern-8pt} & {16} & {32} & {64} & {128} & {16} & {32} & {64} & {128} \\
\midrule
\multirow[c]{3}{*}{\rotatebox{90}{\itot}} & TDSRDH & 78.27 \pm 0.34 & 80.31 \pm 0.19 & 81.81 \pm 0.04 & 82.30 \pm 0.38 & 70.67 \pm 0.33 & 72.60 \pm 0.17 & 73.30 \pm 0.20 & 73.92 \pm 0.30 \\
 & SCH & 83.73 \pm 0.83 & 85.55 \pm 0.23 & 86.66 \pm 0.04 & 88.06 \pm 0.14 & 71.89 \pm 0.68 & 74.12 \pm 0.35 & 75.89 \pm 0.10 & 76.89 \pm 0.13 \\
 & DSCH & \boldmath 84.19 \pm 0.37 & \boldmath 86.49 \pm 0.29 & \boldmath 87.92 \pm 0.14 & \boldmath 88.69 \pm 0.10 & \boldmath 73.64 \pm 0.16 & \boldmath 75.60 \pm 0.12 & \boldmath 76.96 \pm 0.08 & \boldmath 78.03 \pm 0.06 \\
\cmidrule{1-10}
\multirow[c]{3}{*}{\rotatebox{90}{\ttoi}} & TDSRDH & 73.05 \pm 0.34 & 74.79 \pm 0.26 & 76.01 \pm 0.05 & 76.28 \pm 0.44 & 65.23 \pm 0.44 & 67.12 \pm 0.14 & 67.76 \pm 0.13 & 68.37 \pm 0.44 \\
 & SCH & 74.28 \pm 0.61 & 75.96 \pm 0.43 & 77.33 \pm 0.13 & 78.93 \pm 0.02 & 64.32 \pm 0.54 & 66.09 \pm 0.47 & 68.05 \pm 0.30 & 69.10 \pm 0.24 \\
 & DSCH & \boldmath 74.67 \pm 0.07 & \boldmath 76.80 \pm 0.18 & \boldmath 78.62 \pm 0.03 & \boldmath 79.62 \pm 0.23 & \boldmath 65.93 \pm 0.16 & \boldmath 67.79 \pm 0.33 & \boldmath 69.08 \pm 0.22 & \boldmath 70.13 \pm 0.22 \\
\cmidrule{1-10}
\multirow[c]{3}{*}{\rotatebox{90}{\itoi}} & TDSRDH & 73.63 \pm 0.35 & 75.31 \pm 0.24 & 76.55 \pm 0.29 & 76.73 \pm 0.30 & 66.14 \pm 0.56 & 68.04 \pm 0.08 & 68.80 \pm 0.09 & 69.39 \pm 0.25 \\
 & SCH & \boldmath 76.67 \pm 0.68 & 78.31 \pm 0.21 & 79.65 \pm 0.09 & 81.17 \pm 0.09 & 66.21 \pm 0.71 & 68.09 \pm 0.31 & 69.81 \pm 0.25 & 70.96 \pm 0.21 \\
 & DSCH & 76.42 \pm 0.17 & \boldmath 78.52 \pm 0.09 & \boldmath 80.34 \pm 0.14 & \boldmath 81.31 \pm 0.14 & \boldmath 67.78 \pm 0.39 & \boldmath 69.24 \pm 0.16 & \boldmath 70.91 \pm 0.27 & \boldmath 71.92 \pm 0.05 \\
\cmidrule{1-10}
\multirow[c]{3}{*}{\rotatebox{90}{\ttot}} & TDSRDH & \boldmath 69.34 \pm 0.36 & 70.73 \pm 0.19 & 71.58 \pm 0.14 & 70.67 \pm 1.00 & 61.43 \pm 0.57 & \boldmath 63.19 \pm 0.07 & 63.74 \pm 0.15 & 64.13 \pm 0.60 \\
 & SCH & 69.16 \pm 0.59 & \boldmath 70.80 \pm 0.35 & 72.16 \pm 0.19 & 73.52 \pm 0.18 & 60.25 \pm 0.50 & 62.03 \pm 0.34 & 63.66 \pm 0.33 & 64.60 \pm 0.21 \\
 & DSCH & 68.96 \pm 0.02 & \boldmath 70.80 \pm 0.25 & \boldmath 72.69 \pm 0.15 & \boldmath 73.79 \pm 0.37 & \boldmath 61.53 \pm 0.28 & 63.03 \pm 0.33 & \boldmath 64.52 \pm 0.26 & \boldmath 65.36 \pm 0.27 \\

\bottomrule
\end{tabular}%
\end{table*}
\begin{table*}
    \centering
    \robustify\boldmath
\sisetup{
    detect-weight=true,
    table-format=2.2(2),
    table-model-setup=\boldmath,
    table-alignment-mode=format,
    table-number-alignment=left,
    table-text-alignment=center,
    reset-math-version=false,
}
\caption{Comparison of $\gamma_w$ hyperparameter by tie-aware mean average precision ($\text{mAP}_\text{T}$) scores (in percent) for four tasks on the validation set. Hash code length is $k=32$ for all experiments. Top-3 performances per task and dataset are highlighted in boldface.}
\label{t:e2:map}
\begin{tabular}{@{\hspace{3pt}}l@{\hspace{5pt}}l@{\hspace{8pt}}S*{0}{@{\hspace{5pt}}S}@{\hspace{12pt}}S*{0}{@{\hspace{5pt}}S}@{\hspace{12pt}}S*{0}{@{\hspace{5pt}}S}@{\hspace{12pt}}S*{0}{@{\hspace{5pt}}S}@{\hspace{12pt}}S*{0}{@{\hspace{5pt}}S}@{\hspace{12pt}}S*{0}{@{\hspace{5pt}}S}@{\hspace{12pt}}S*{0}{@{\hspace{5pt}}S}@{\hspace{3pt}}}
\toprule
{\multirow[b]{2}{*}{\rotatebox{90}{Task}}} & {$\gamma_w$} & {2} & {4} & {6} & {8} & {10} & {12} & {14} \\
{} & {Dataset} & {} & {} & {} & {} & {} & {} & {} \\
\midrule
\multirow[c]{2}{*}{\rotatebox{90}{\itot}} & \MIRFLICKR & 83.67 \pm 0.17 & 84.25 \pm 0.27 & 84.41 \pm 0.10 & \boldmath 84.53 \pm 0.44 & 84.16 \pm 0.14 & \boldmath 84.57 \pm 0.12 & \boldmath 84.43 \pm 0.30 \\
 & \NUSWIDE & 75.04 \pm 0.08 & 75.69 \pm 0.17 & \boldmath 75.92 \pm 0.09 & 75.83 \pm 0.17 & \boldmath 75.92 \pm 0.20 & 75.91 \pm 0.14 & \boldmath 75.95 \pm 0.13 \\
\cmidrule{1-9}
\multirow[c]{2}{*}{\rotatebox{90}{\ttoi}} & \MIRFLICKR & 74.67 \pm 0.07 & \boldmath 75.19 \pm 0.21 & \boldmath 75.22 \pm 0.33 & \boldmath 75.22 \pm 0.23 & 74.98 \pm 0.15 & 75.15 \pm 0.08 & 75.16 \pm 0.37 \\
 & \NUSWIDE & 66.90 \pm 0.15 & 67.67 \pm 0.30 & \boldmath 67.92 \pm 0.05 & \boldmath 67.85 \pm 0.39 & 67.84 \pm 0.24 & \boldmath 67.93 \pm 0.19 & 67.81 \pm 0.12 \\
\cmidrule{1-9}
\multirow[c]{2}{*}{\rotatebox{90}{\itoi}} & \MIRFLICKR & 75.86 \pm 0.19 & 76.53 \pm 0.43 & \boldmath 76.64 \pm 0.09 & \boldmath 76.86 \pm 0.05 & 76.21 \pm 0.27 & 76.60 \pm 0.00 & \boldmath 76.76 \pm 0.30 \\
 & \NUSWIDE & 69.03 \pm 0.18 & 69.94 \pm 0.35 & \boldmath 70.20 \pm 0.28 & \boldmath 70.14 \pm 0.21 & 70.02 \pm 0.17 & 70.13 \pm 0.57 & \boldmath 70.21 \pm 0.16 \\
\cmidrule{1-9}
\multirow[c]{2}{*}{\rotatebox{90}{\ttot}} & \MIRFLICKR & 68.84 \pm 0.06 & 69.46 \pm 0.31 & \boldmath 69.52 \pm 0.23 & \boldmath 69.60 \pm 0.08 & 69.29 \pm 0.10 & 69.45 \pm 0.05 & \boldmath 69.67 \pm 0.33 \\
 & \NUSWIDE & 62.45 \pm 0.35 & 63.43 \pm 0.41 & \boldmath 63.60 \pm 0.20 & \boldmath 63.65 \pm 0.38 & 63.51 \pm 0.17 & \boldmath 63.59 \pm 0.51 & \boldmath 63.59 \pm 0.15 \\

\bottomrule
\end{tabular}%
\end{table*}
\begin{table*}
    \centering
    \robustify\boldmath
\sisetup{
    detect-weight=true,
    table-format=2.2(2),
    table-model-setup=\boldmath,
    table-alignment-mode=format,
    table-number-alignment=left,
    table-text-alignment=center,
    reset-math-version=false,
}
\caption{Comparison of $\gamma_w$ hyperparameter by ROC-AUC scores (in percent) for four tasks on the validation set. Hash code length is $k=32$ for all experiments. Best performance per task and dataset is highlighted in boldface.}
\label{t:e2:ra}
\begin{tabular}{@{\hspace{3pt}}l@{\hspace{5pt}}l@{\hspace{8pt}}S*{0}{@{\hspace{5pt}}S}@{\hspace{12pt}}S*{0}{@{\hspace{5pt}}S}@{\hspace{12pt}}S*{0}{@{\hspace{5pt}}S}@{\hspace{12pt}}S*{0}{@{\hspace{5pt}}S}@{\hspace{12pt}}S*{0}{@{\hspace{5pt}}S}@{\hspace{12pt}}S*{0}{@{\hspace{5pt}}S}@{\hspace{12pt}}S*{0}{@{\hspace{5pt}}S}@{\hspace{3pt}}}
\toprule
{\multirow[b]{2}{*}{\rotatebox{90}{Task}}} & {$\gamma_w$} & {2} & {4} & {6} & {8} & {10} & {12} & {14} \\
{} & {Dataset} & {} & {} & {} & {} & {} & {} & {} \\
\midrule
\multirow[c]{2}{*}{\rotatebox{90}{\itot}} & \MIRFLICKR & 72.15 \pm 0.04 & 72.90 \pm 0.44 & 73.03 \pm 0.16 & \boldmath 73.40 \pm 0.24 & 73.05 \pm 0.49 & 73.24 \pm 0.19 & 73.30 \pm 0.30 \\
 & \NUSWIDE & 79.58 \pm 0.14 & 80.94 \pm 0.33 & 81.43 \pm 0.09 & \boldmath 81.55 \pm 0.17 & 81.28 \pm 0.32 & 81.45 \pm 0.13 & 81.40 \pm 0.10 \\
\cmidrule{1-9}
\multirow[c]{2}{*}{\rotatebox{90}{\ttoi}} & \MIRFLICKR & 71.50 \pm 0.23 & 72.25 \pm 0.32 & 72.31 \pm 0.19 & \boldmath 72.57 \pm 0.25 & 72.08 \pm 0.45 & 72.45 \pm 0.15 & \boldmath 72.57 \pm 0.28 \\
 & \NUSWIDE & 79.91 \pm 0.14 & 81.20 \pm 0.30 & 81.64 \pm 0.24 & \boldmath 81.78 \pm 0.39 & 81.36 \pm 0.16 & 81.66 \pm 0.30 & 81.74 \pm 0.11 \\
\cmidrule{1-9}
\multirow[c]{2}{*}{\rotatebox{90}{\itoi}} & \MIRFLICKR & 80.76 \pm 0.33 & 81.40 \pm 0.37 & 81.81 \pm 0.07 & 81.94 \pm 0.57 & 81.68 \pm 0.13 & \boldmath 82.08 \pm 0.30 & 81.89 \pm 0.24 \\
 & \NUSWIDE & 85.21 \pm 0.41 & 86.36 \pm 0.18 & 86.89 \pm 0.13 & \boldmath 87.12 \pm 0.08 & 86.96 \pm 0.20 & 86.93 \pm 0.03 & 86.93 \pm 0.11 \\
\cmidrule{1-9}
\multirow[c]{2}{*}{\rotatebox{90}{\ttot}} & \MIRFLICKR & 64.71 \pm 0.16 & 65.60 \pm 0.41 & 65.52 \pm 0.25 & \boldmath 65.94 \pm 0.04 & 65.57 \pm 0.45 & 65.67 \pm 0.09 & 65.84 \pm 0.20 \\
 & \NUSWIDE & 74.96 \pm 0.35 & 76.45 \pm 0.38 & 76.80 \pm 0.23 & 76.91 \pm 0.33 & 76.55 \pm 0.34 & 76.82 \pm 0.42 & \boldmath 76.92 \pm 0.11 \\

\bottomrule
\end{tabular}%
\end{table*}
\begin{table*}
    \centering
    \robustify\boldmath
\sisetup{
    detect-weight=true,
    table-format=2.2(2),
    table-model-setup=\boldmath,
    table-alignment-mode=format,
    table-number-alignment=left,
    table-text-alignment=center,
    reset-math-version=false,
}
\caption{Comparison of three loss functions by tie-aware mean average precision ($\text{mAP}_\text{T}$) scores (in percent) for four tasks on the test set. The hash code length is $k=32$ for all experiments. Best performance per model, dataset, and task is highlighted in boldface.}
\label{t:e3:map}
\begin{tabular}{@{\hspace{3pt}}l@{\hspace{5pt}}l@{\hspace{8pt}}S*{1}{@{\hspace{5pt}}S}@{\hspace{12pt}}S*{1}{@{\hspace{5pt}}S}@{\hspace{3pt}}}
\toprule
{\multirow[b]{3}{*}{\rotatebox{90}{Task}}} & {Model} & \multicolumn{2}{c}{\CLIPHash} & \multicolumn{2}{c}{\TDSRDH} \\
{} & {Dataset} & {\MIRFLICKR} & {\NUSWIDE} & {\MIRFLICKR} & {\NUSWIDE} \\
{} & {Loss} & {} & {} & {} & {} \\
\midrule
\multirow[c]{3}{*}{\rotatebox{90}{\itot}} & TDSRDH & 86.36 \pm 0.03 & 74.52 \pm 0.22 & 80.31 \pm 0.19 & 72.60 \pm 0.17 \\
 & SCH & 88.82 \pm 0.50 & 76.53 \pm 0.32 & 85.55 \pm 0.23 & 74.12 \pm 0.35 \\
 & DSCH & \boldmath 89.51 \pm 0.32 & \boldmath 77.12 \pm 0.15 & \boldmath 86.49 \pm 0.29 & \boldmath 75.60 \pm 0.12 \\
\cmidrule{1-6}
\multirow[c]{3}{*}{\rotatebox{90}{\ttoi}} & TDSRDH & 81.93 \pm 0.09 & 71.17 \pm 0.20 & 74.79 \pm 0.26 & 67.12 \pm 0.14 \\
 & SCH & 82.12 \pm 0.48 & 71.30 \pm 0.26 & 75.96 \pm 0.43 & 66.09 \pm 0.47 \\
 & DSCH & \boldmath 83.25 \pm 0.24 & \boldmath 72.17 \pm 0.16 & \boldmath 76.80 \pm 0.18 & \boldmath 67.79 \pm 0.33 \\
\cmidrule{1-6}
\multirow[c]{3}{*}{\rotatebox{90}{\itoi}} & TDSRDH & 81.92 \pm 0.13 & 71.79 \pm 0.17 & 75.31 \pm 0.24 & 68.04 \pm 0.08 \\
 & SCH & 82.63 \pm 0.48 & 72.47 \pm 0.40 & 78.31 \pm 0.21 & 68.09 \pm 0.31 \\
 & DSCH & \boldmath 83.27 \pm 0.29 & \boldmath 73.10 \pm 0.15 & \boldmath 78.52 \pm 0.09 & \boldmath 69.24 \pm 0.16 \\
\cmidrule{1-6}
\multirow[c]{3}{*}{\rotatebox{90}{\ttot}} & TDSRDH & \boldmath 78.05 \pm 0.08 & \boldmath 68.68 \pm 0.15 & 70.73 \pm 0.19 & \boldmath 63.19 \pm 0.07 \\
 & SCH & 76.88 \pm 0.45 & 67.84 \pm 0.32 & \boldmath 70.80 \pm 0.35 & 62.03 \pm 0.34 \\
 & DSCH & 77.83 \pm 0.18 & 68.66 \pm 0.12 & \boldmath 70.80 \pm 0.25 & 63.03 \pm 0.33 \\

\bottomrule
\end{tabular}%
\end{table*}

\section{Conclusion and Future Work}\label{s:conc}
In this paper, we have proposed a new loss function in the field of \acl{dsh} for the tasks of cross-modal as well as intra-modal retrieval. The proposed loss function, \acl{dsch}, extends prior work by providing a continuous loss landscape and a dynamic target sematic channel width in the shared Hamming space dependent on label similarity. The efficacy of \ac{dsch} was demonstrated in various experimental settings comparing our method with prior work on both an existing dual-stream model architecture and a hybrid-stream model architecture, as well as on two datasets commonly found in literature. These experiments showed significantly better tie-aware \acl{map} scores for all cross-modal tasks as well as the intra-modal retrieval using images. For larger hash code lengths, \ac{dsch} outperforms its competitors in all four tasks. The metric of tie-aware \ac{map} scores was chosen to avoid evaluation bias emerging from the ambiguity of retrieval ordering, which is inherent to the \ac{dsh} domain due to a limited amount of discrete sample-to-sample distances in the shared Hamming space.

Nonetheless, there are still many aspects in the \ac{dsh} domain subject to further research. For example, the contrastive learning approach employed in the pre-trained \CLIPHash model used in the experiments may be transferred to the \ac{dsh} domain. Albeit the better performance of the \CLIPHash model over the \TDSRDH model architecture in the conducted experiments, the former is much larger in terms of parameters and number of layers and, thus, has vastly higher inference costs in both compute time and memory. Distillation learning in a teacher-student setting might provide remedy and allow smaller models to achieve similar scores at lower resource consumption. Additionally, the training procedure can be enriched with a wider set of data augmentation techniques, especially for the text domain, which was left untouched by the experiments in this paper. For example, the effect of token masking and random token replacement could be studied. The image data augmentation pipeline can be extended as well, e.g., with more multi-image non-instance methods involving more than two images. This approach may also include the associated sample labels, which could be fused and/or mixed alongside the images or texts during data augmentation. Future work may look into the generation of adversarial examples during the training procedure for hard negative mining. Another aspect is the analysis of hash code generation robustness and trustworthiness, e.g., by applying explainable AI techniques, and the development of incremental hashing methods.

\renewcommand*{\bibfont}{\footnotesize}
\printbibliography
\end{document}